\documentclass[letterpaper]{article} 
\usepackage{aaai2026}  
\usepackage{times}  
\usepackage{helvet}  
\usepackage{courier}  
\usepackage[hyphens]{url}  
\usepackage{graphicx} 
\usepackage{natbib}  
\usepackage{caption} 
\usepackage{algorithm}
\usepackage{algorithmic}
\usepackage{booktabs} 
\usepackage{amsmath} 
\usepackage{amsfonts}
\usepackage{utfsym}

\usepackage{amssymb}
\usepackage{colortbl}
\usepackage{xcolor}
\usepackage{natbib}
\usepackage{bbding}
\usepackage{color}
\usepackage{colortbl}

\definecolor{Red}{cmyk}{0,1,1,0}
\definecolor{Green}{cmyk}{1,0,1,0}
\definecolor{Cyan}{cmyk}{1,0,0,0}
\definecolor{Purple}{cmyk}{0.45,0.86,0,0}
\definecolor{Rosolic}{cmyk}{0.00,1.00,0.50,0}
\definecolor{Blue}{cmyk}{1.00,1.00,0.00,0}
\definecolor{Orange}{cmyk}{0,0.52,0.80,0}
\definecolor{Black}{cmyk}{1,0,0,1}

\newcommand{\lyj}{}

\newcommand{\yh}{}
\usepackage{newfloat}
\usepackage{listings}
\DeclareCaptionStyle{ruled}{labelfont=normalfont,labelsep=colon,strut=off} 
\floatstyle{ruled}
\newfloat{listing}{tb}{lst}{}
\floatname{listing}{Listing}
\title{S\textsuperscript{3}LAM: Surfel Splatting SLAM for Geometrically Accurate \\ Tracking and Mapping}
\author{
    Ruoyu Fan\textsuperscript{\rm 1}, Yuhui Wen\textsuperscript{\rm 2}, Jiajia Dai\textsuperscript{\rm 1}, Tao Zhang\textsuperscript{\rm 3}, Long Zeng\textsuperscript{\rm 4}, Yong-jin Liu\textsuperscript{\rm 1}
}
\affiliations{
    \textsuperscript{\rm 1}Department of Computer Science and Technology, Tsinghua University\\
    \textsuperscript{\rm 2}School of Computer Science and Technology, Beijing Jiaotong University\\
    \textsuperscript{\rm 3}Pudu Robotics, 
    \textsuperscript{\rm 4}Tsinghua Shenzhen International Graduate School\\

}

\usepackage{bibentry}

\begin{document}

\maketitle

\begin{abstract}
We propose S\textsuperscript{3}LAM, a \lyj{novel RGB-D} SLAM system \lyj{that leverages 2D} surfel splatting to achieve highly accurate \lyj{geometric} representations for simultaneous tracking and mapping. Unlike existing \lyj{3DGS-based} SLAM \lyj{approaches that rely on 3D Gaussian ellipsoids}, we utilize 2D Gaussian surfels as primitives for \lyj{more efficient} scene representation. 
By focusing on the surfaces of objects in the scene, this design enables S\textsuperscript{3}LAM to reconstruct high-quality geometry, benefiting both mapping and tracking.
To address inherent SLAM challenges including \lyj{real-time} optimization under \lyj{limited} viewpoints, we introduce a \lyj{novel} adaptive surface rendering \lyj{strategy} that \lyj{improves} mapping accuracy while maintaining computational efficiency. \lyj{We further} derive \lyj{camera} pose Jacobians directly from \lyj{2D} surfel splatting formulation, 
highlighting the importance of our geometrically accurate representation \lyj{that improves} tracking convergence. 
Extensive experiments on \lyj{both} synthetic and real-world datasets \lyj{validate that S\textsuperscript{3}LAM achieves} state-of-the-art performance. \lyj{Code} will be \lyj{made} publicly \lyj{available}.
\end{abstract}


\section{Introduction}

Simultaneous Localization and Mapping (SLAM) is a crucial component in various applications, \lyj{such as} autonomous driving, mobile robotics, and augmented reality. Over the past few decades, SLAM research has explored various representations for scene reconstruction, including point clouds~\cite{monoslam}, meshes~\cite{ovpcmesh}, and voxel grids~\cite{dai2017bundlefusion, kinectfusion}, which have achieved \lyj{good} tracking and mapping performance. 

Recently, representing scenes with neural radiance fields~(NeRF) ~\cite{mildenhall2021nerf} has gained significant attention \lyj{in SLAM research}~\cite{sucar2021imap, yang2022vox}. \lyj{These} approaches combine view rendering and surface reconstruction 
with dense visual SLAM, expanding the potential applications \lyj{of} SLAM systems.
\lyj{Along this direction,} a growing trend in visual SLAM is the use of 3D Gaussian Splatting (3DGS)~\cite{kerbl3Dgaussians} 
for scene mapping. Compared to radiance fields, 3DGS offers superior rendering speed,  
enhanced interpretability and flexible extensibility. \lyj{3DGS-based} SLAM systems effectively integrate \lyj{3D} Gaussian splatting techniques, incorporating innovations in Gaussian management~\cite{yan2024gs}, pose optimization~\cite{matsuki2024gaussian}, and uncertainty estimation~\cite{hu2025cg}. These \lyj{advancements demonstrate the significant potential of 3DGS for SLAM applications}. 

\begin{figure}[tbp]\centering
  \includegraphics[width=1.0\linewidth]{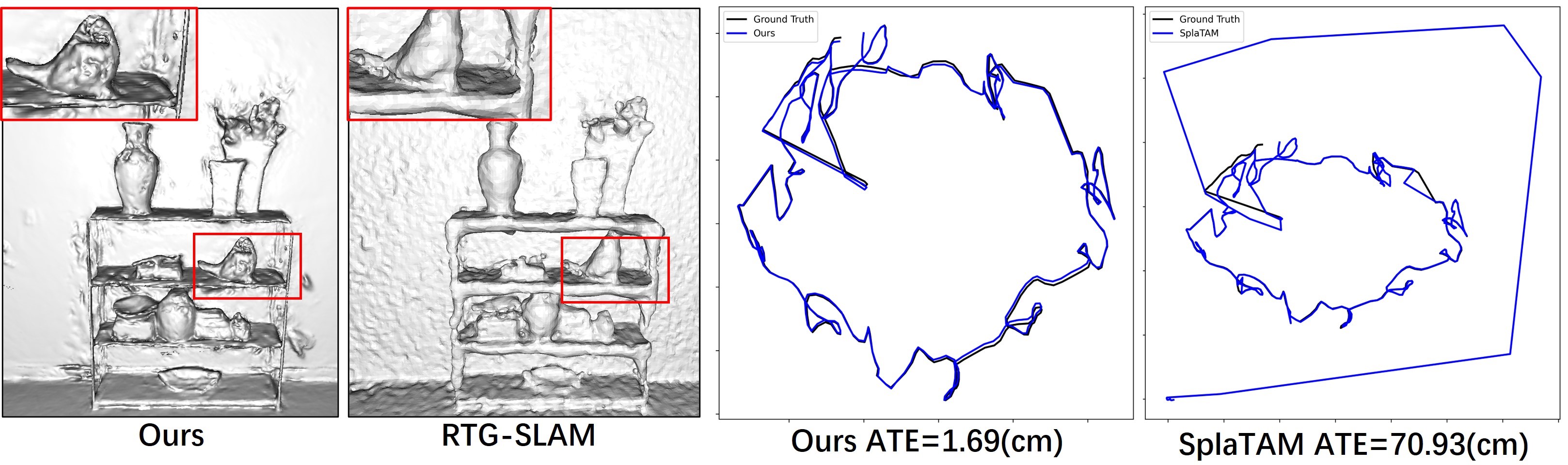}
    \caption{Compared to state-of-the-art 3DGS-based SLAM systems~\cite{peng2024rtg, keetha2024splatam}, our method achieves geometrically accurate scene reconstruction with fine details (Left vs. RTG-SLAM on {\it Room2}, Replica) and improved convergence in camera pose tracking (Right vs. SplaTAM on {\it S1}, ScanNet++).}
  \label{fig:title}
\end{figure}

Our work is motivated by the following observation: 3DGS-based SLAM methods reconstruct scenes \lyj{using inverse rendering} that prioritizes color consistency, which often \lyj{compromises} the reconstruction of \lyj{fine geometric details} such as surfaces. 
Since tracking and mapping are tightly coupled in SLAM, degraded mapping representation can in turn \lyj{deteriorate} tracking performance. 
The inherent \lyj{characteristics} of \lyj{3DGS} constrain \lyj{the SLAM performance} in both mapping and tracking, as \lyj{summarized} below:

\begin{itemize}
\item \textbf{Mapping}: \lyj{3DGS reconstructs scene geometry by fitting the average depth of projected ellipsoids. Due to the lack of explicit surface normals, it struggles to accurately represent object surfaces, leading to visual artifacts and geometric inconsistencies across views.}
\item \textbf{Tracking}: \lyj{Similarly, the absence of explicit surface normals in 3DGS restricts camera tracking to only optimize the value of Gaussian distribution functions, making it difficult for the estimated camera rotations to align with the true surface orientations in the scene.}
\end{itemize}

To \lyj{overcome the above} limitations, \lyj{in this paper,} we adopt \lyj{2D} Gaussian surfel splatting as scene representation primitives, inspired by \lyj{the recent success of} offline 2D Gaussian splatting approaches~\cite{huang20242d, dai2024high}. \lyj{For} SLAM \lyj{applications} with limited Gaussian primitives, we introduce an adaptive surface reconstruction strategy that dynamically refines geometry. \lyj{Furthermore}, we \lyj{present} a novel pose estimation \lyj{algorithm that explicitly leverages 2D} surfels to robustly handle extreme viewpoint changes.

\lyj{To sum up,} we propose {\it Surfel Splatting SLAM} (S\textsuperscript{3}LAM), a novel system designed to achieve geometrically accurate mapping and tracking \lyj{through 2D Gaussian surfel representations}. \lyj{S\textsuperscript{3}LAM} integrates forward rendering for scene reconstruction with backward gradient propagation to \lyj{jointly} optimize color, depth, and normal \lyj{estimation}. 
To enable online adaptive mapping, we first quantify the uncertainty in different regions by introducing a depth distortion term during optimization. This distortion term identifies areas \lyj{under construction} in mapping, \lyj{triggering our} adaptive reconstruction strategy that replaces uncertain geometry with dominant surfels. 
\lyj{For camera tracking, we derive an analytic Jacobian on Lie algebra for the surfel splatting map. Our approach incorporates a radial gradient in the surfel-based Jacobian, enabling explicit alignment optimization between the camera orientation and reconstructed surfaces.}
These designs enable our method to achieve accurate geometric reconstruction and robust tracking convergence (Figure~\ref{fig:title}).

Our contributions \lyj{made in this paper include:} 
\begin{itemize}
    \item We propose S\textsuperscript{3}LAM, a \lyj{novel} real-time SLAM system \lyj{that utilizes} 2D Gaussian surfel primitives \lyj{to achieve both accurate and efficient scene representation.} 
    \item We introduce an online adaptive mapping strategy \lyj{coupled with a novel} surfel-based pose estimation \lyj{algorithm}, \lyj{significantly improving tracking and mapping performance in SLAM} under challenging conditions \lyj{such as severe viewpoint changes}. 
    \item We validate the effectiveness of surfel-based tracking through convergence basin analysis. S\textsuperscript{3}LAM demonstrates state-of-the-art performance in both mapping and tracking compared to NeRF-based and 3DGS-based SLAM systems on multiple datasets.
\end{itemize}


\begin{figure*}[htbp]\centering
  \includegraphics[width=0.95\linewidth]{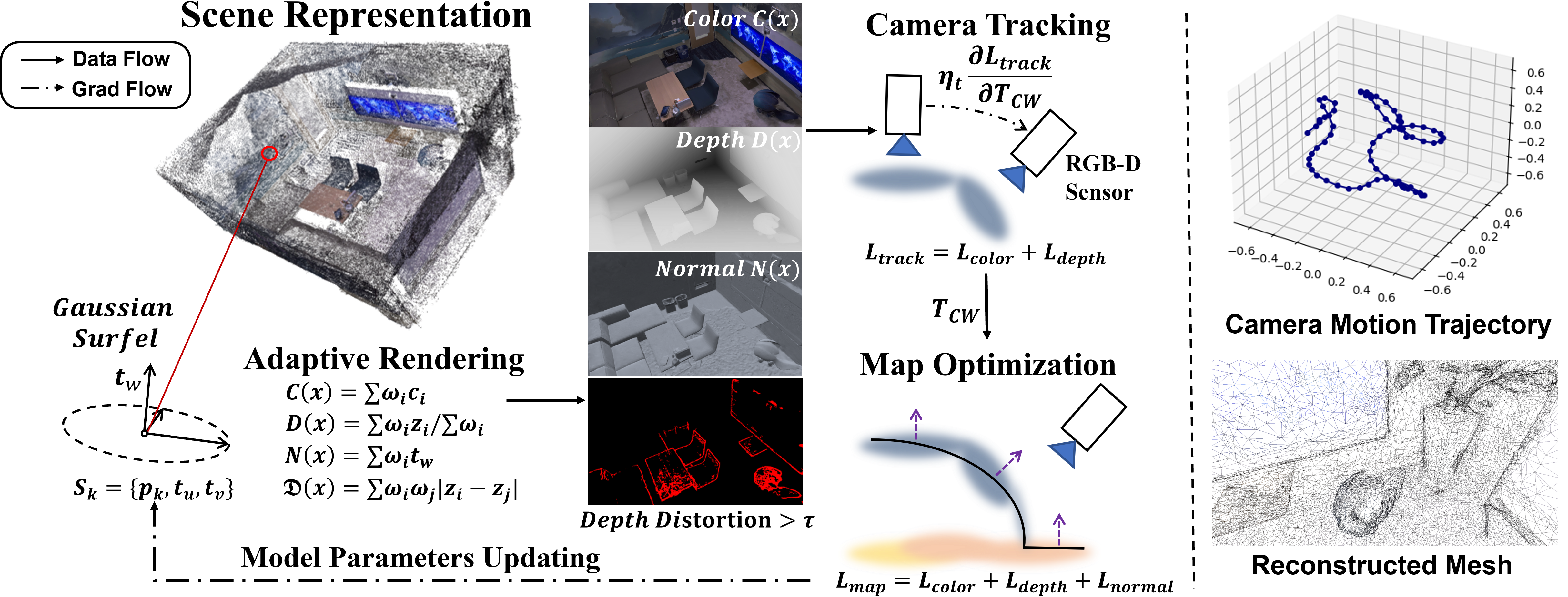}
    \caption{\textbf{Overview of the proposed S\textsuperscript{3}LAM System.} The scene is represented \lyj{by 2D} Gaussian surfels to achieve geometry-aligned motion tracking and scene mapping. Both the tracking and mapping phases benefit from our oriented surfel primitives, leveraging the proposed adaptive rendering and surfel-based pose estimation. Our system outputs the motion trajectory and a reconstructed mesh model of the scene after post-processing.}
  \label{fig:method}
\end{figure*}

\section{Related Work}

{\bf Dense Visual SLAM} \lyj{aims} 
to achieve simultaneous localization and mapping \lyj{through} reconstructing dense 3D scene representations. Early breakthroughs, \lyj{e.g.,} DTAM~\cite{newcombe2011dtam} and KinectFusion~\cite{kinectfusion}, 
laid the foundation for dense SLAM by leveraging photometric consistency and TSDF fusion, respectively, to represent and update the scene. Subsequent works, \lyj{such as} ElasticFusion~\cite{whelan2016elasticfusion} and DI-Fusion~\cite{huang2021di}, explored a wide range of map representations, including point clouds, surfels, voxel hashing and octrees, to tackle challenges in scalability and accuracy. 
In parallel, recent advancements have integrated deep learning into traditional frameworks, \lyj{e.g.,} SceneCode~\cite{zhi2019scenecode}, NodeSLAM~\cite{sucar2020nodeslam} and DROID-SLAM~\cite{teed2021droid}, to achieve robust camera tracking and mapping. These developments significantly expanded the applications of dense SLAM systems. 

{\bf NeRF-based SLAM} \lyj{has} emerged as a powerful approach for dense visual SLAM, leveraging the advancements in neural radiance fields (NeRF), to jointly optimize scene representation and camera poses. 
iMAP~\cite{sucar2021imap} introduced single-MLP-based scene representation for scalable mapping. 
NICE-SLAM~\cite{zhu2022nice} and Vox-Fusion~\cite{yang2022vox} incorporated hierarchical voxel grids and octree structures, 
to enhance scalability and precision. 
\yh{Recently,} Point-SLAM~\cite{sandstrom2023point} \lyj{achieved detailed reconstruction by using} neural point clouds with volumetric rendering. \lyj{However,} due to the computational requirements of volume rendering, \lyj{it faces} challenges in real-time performance. 
\lyj{Although} these methods have demonstrated remarkable accuracy, their reliance on memory-intensive structures and time-consuming rendering limits their scalability and real-time applicability. 

{\bf 3DGS-based SLAM.} Recent advancements have integrated 3D Gaussians into \lyj{RGB-D} dense SLAM systems, 
to represent and render high-quality scenes. 
\yh{The pioneering work} \cite{kerbl3Dgaussians} demonstrated the effectiveness of 3D Gaussians in photorealistic real-time rendering, but their optimization processes were offline and computationally expensive. To adapt Gaussians for online reconstruction, \lyj{Yan et al.}~\shortcite{yan2024gs} introduced adaptive expansion and coarse-to-fine tracking. SplaTAM~\cite{keetha2024splatam} tailored a pipeline to optimize Gaussians with silhouette-guided differentiable rendering. 
\lyj{Matsuki et al.}~\shortcite{matsuki2024gaussian} presented isotropic Gaussian-based tracking, mapping, and rendering for RGBD SLAM. \lyj{Despite} these innovations, real-time performance \lyj{in} large-scale scenes \lyj{remains challenging}. 
RTG-SLAM~\cite{peng2024rtg} sought to unify Gaussian splats for real-time tracking and mapping. 
A concurrent work GauS-SLAM~\cite{su2025gausslamdensergbdslam} also adopted 2D Gaussian surfels as the representation in SLAM. However, their focus lies primarily on local mapping and surfel management, whereas our method emphasizes high-certainty surface reconstruction and surfel-based pose tracking with strong convergence properties.

We begin by briefly summarizing the 2D Gaussian surfel splatting technique, followed by a detailed presentation of our S\textsuperscript{3}LAM system.

\section{Preliminary on 2D Gaussian Surfel Splatting}
\label{surfel}

\lyj{Our S\textsuperscript{3}LAM system adopts 2D}  Gaussian surfels \cite{huang20242d} to represent the scenes, \lyj{in which} surfel splatting is \lyj{performed by} blending Gaussian surfels intersected by rays \lyj{from image pixels}. A surfel is characterized by its center $p_k$, along with two principal tangential vectors $\mathbf{t}_u$ and $\mathbf{t}_v$, and 
\yh{their corresponding} scaling factors $s_u$ and $s_v$. The normal vector is defined by the cross product $\mathbf{t}_w = \mathbf{t}_u \times \mathbf{t}_v$, and the rotation matrix is represented as $\mathbf{R} = [\mathbf{t}_u, \mathbf{t}_v, \mathbf{t}_w]$. A local point on a Gaussian surfel is parameterized by the $(u, v)$ coordinates:
\begin{small}    
\begin{equation}
P_k(u, v) = p_k + s_u \mathbf{t}_u u + s_v \mathbf{t}_v v
\end{equation}
\end{small}

The local coordinates $(u, v)$ \lyj{are} computed by solving the equation $ z(x, y, 1) = [\mathbf{R} \,|\, \mathbf{t}] P_k(u, v) $, where $z(x, y, 1)$ is the point in the camera coordinate system, \yh{and} $[\mathbf{R} \,|\, \mathbf{t}]$ represents the transformation matrix consisting of the rotation matrix $\mathbf{R}$ and translation vector $\mathbf{t}$. Given the local coordinates, the ray-surfel transparency \yh{$\omega_{i}$ }
is calculated as follows:
\begin{small}
\begin{align}
&G_k(\mathbf{u}(\mathbf{x})) = \exp(-u^2/2 - v^2/2) \\
&\omega_{i} = \alpha_i \, G_i(\mathbf{u}(\mathbf{x})) \prod_{j=1}^{i-1} \left( 1 - \alpha_j G_j(\mathbf{u}(\mathbf{x})) \right)
\label{eqn:01}
\end{align}
\end{small}

Here, \(\mathbf{u}(\mathbf{x})\) denotes the local coordinates \((u, v)\) determined by the intersection between the pixel ray from \(\mathbf{x}\) and the surfel plane. 
$\alpha_i$ is the inherent opacity of the $i$-th surfel. 
Color, depth, and normal values are rendered using alpha blending, which is computed as the weighted sum of alpha weights $\omega_k$ in a front-to-back order of Gaussian surfels: 
\begin{scriptsize}
\begin{equation}
\label{equ:render}
\mathbf{C}(\mathbf{x}) = \sum_{k=1}^{K} \omega_k \, c_k, \, 
\mathbf{D}(\mathbf{x}) = \sum_{k=1}^{K} \omega_k \, z_k / \sum_{k=1}^{K} \omega_k , \, 
\mathbf{N}(\mathbf{x}) = \sum_{k=1}^{K} \omega_k \, \mathbf{t}_w
\end{equation} 
\end{scriptsize}

\yh{where} $c_k$ represents the color feature vector of the $k$-th surfel, and $z_k$ represents the $z$-coordinate of the world point $P_k(u, v)$ transformed into the camera \yh{coordinate system.} 
\yh{With the Gaussian surfel representation, our map is} denoted as $\mathbf{S} = \{(p_k, \mathbf{t}_u, \mathbf{t}_v, s_u, s_v, c_k, \alpha_k)\}$.

\section{Method}

\lyj{We propose S\textsuperscript{3}LAM (short for \underline{S}urfel \underline{S}platting \underline{SLAM}), a novel RGB-D SLAM system that leverages 2D surfel splatting for robust and efficient scene mapping. This section is organized as follows: (1) we introduce the 2D Gaussian surfel splatting technique for an efficient map representation; (2) then we present our adaptive reconstruction and surfel management strategy using uncertainty measures, and (3) finally we propose a novel pose optimization algorithm by deriving surfel Jacobians, which improves tracking convergence. The overview of the whole S\textsuperscript{3}LAM system is illustrated in Figure~\ref{fig:method}.}


\subsection{Surfel Splatting Mapping}
\label{mapping}

Given the RGB-D input stream $\{\bar{\mathbf{C}}_t, \bar{\mathbf{D}}_t\}_{t=1}^N$, we first compute the local normal map $\bar{N}_t$, 
by calculating the spatial gradients of the depth map \cite{kinectfusion}. 
With the pose $T_{CW}$ estimated by the camera tracking module, the surfels are rendered to \lyj{determine} color ${\mathbf{C}}_t$, depth ${\mathbf{D}}_t$ and normal ${\mathbf{N}}_t$. The mapping optimization loss is defined as a weighted mixture of $L_1$ losses and cosine similarity:
\begin{equation}
L_{map} = \left\lVert \mathbf{C}_t - \mathbf{\bar{C}}_t \right\rVert_1 + \gamma_D\left\lVert \mathbf{D}_t - \mathbf{\bar{D}}_t \right\rVert_1 + \gamma_N\left( 1 - \mathbf{N}_t\cdot\mathbf{\bar{N}}_t \right)
\end{equation}
where $\left\lVert \cdot \right\rVert_1$ denotes $L_1$ loss over image pixels, $\mathbf{N}_t\cdot\mathbf{\bar{N}}_t$ represents the pixel-wise cosine similarity between the groundtruth normal and estimated normal. 

\lyj{{\it Adaptive Surface Mapping.} The original}  surfel splatting \lyj{technique \cite{huang20242d} is offline, which can} 
effectively reconstruct scene geometry under sufficient optimization iterations, \lyj{using} a large number of surfels for representation and \yh{ensuring} complete multiview coverage. However, in SLAM systems, \lyj{the time} constraints \lyj{imposed by real-time} mapping and tracking optimization necessitate \lyj{the use of}
a reduced number of \lyj{surfels} and a \lyj{rapidly} converging pipeline. Under these \lyj{constraints}, conventional surfel splatting often \lyj{results in} incomplete surface reconstruction, \lyj{particularly at sharp geometric features such as edges and corners, thereby compromising the accuracy of surface modeling.} 
\label{adaptive}

To \lyj{introduce 2D surfel splatting into a real-time SLAM system}, we propose an adaptive rendering strategy that \lyj{can be} seamlessly integrated into the surfel rasterization pipeline, leveraging a depth distortion term $\mathcal{D}_d$ to identify the uncertainty of pixel \lyj{rays}. $\mathcal{D}_d$ is defined as:
\begin{equation}
    \mathcal{D}_d = \sum_{i,j} \omega_i \omega_j |z_i - z_j|
\end{equation}
where $\omega_i$ is the same as in \lyj{Eq.~(\ref{eqn:01})}, and $z_i$ is the depth of the $i$-th intersection point. The depth distortion term was originally used to consolidate separate volumes belonging to the same surface~\cite{Barron_Mip_NeRF}.  
\lyj{Given} a limited number of surfels, regions with high depth distortion typically contain multiple surfels exhibiting large variations in depth \(z_i\) and a relatively \lyj{uniform} distribution of blending weights \(\omega_i\). 
In our work, we leverage depth distortion as a separation metric to guide accurate surface reconstruction.

\yh{Based on the above observation}, we define the uncertainty mask as 
\yh{$\mathcal{D}_d > \tau$}, where $\tau$ \lyj{represents} the distortion threshold. For pixels with uncertainty \lyj{exceeding} $\tau$, depth and normal values are \lyj{determined not through} \yh{opacity-weighted averages,} 
\lyj{but rather by selecting} the surfel with the maximum blending weight. 
The depth and normal \yh{values} are then directly determined by the intersection of the ray with the plane defined by this surfel: 
\begin{equation}
\mathbf{D}_c(\mathbf x) = \mathop{\arg\max}\limits_{\omega_k} z_k, \ 
\mathbf{N}_c(\mathbf x) = \mathop{\arg\max}\limits_{\omega_k} \mathbf{t}_w
\end{equation}
\lyj{It is worth noting that} the degradation of Gaussian surfels into concrete surfels may inadvertently \lyj{affect} large planar surfaces, potentially resulting in artifacts such as holes on the reconstructed surfaces.
To \lyj{address this issue}, we \lyj{selectively replace} the rendered depth \(\mathbf{D}\) and normal \(\mathbf{N}\) with \(\mathbf{D}_c\) and \(\mathbf{N}_c\) only when \(\mathbf{D}(\mathbf{x}) > \mathbf{D}_c(\mathbf{x})\). This \lyj{conditional substitution} ensures that \lyj{our} adaptive mapping strategy preserves surface integrity while effectively managing uncertainty.

Our adaptive rendering strategy is readily integrated into the surfel rasterization pipeline in both the forward and backward passes, requiring only the retention of dominant surfel information during pixel rasterization. With negligible computational overhead, our strategy simultaneously improves sharp geometry representation and effectively identifies regions of high uncertainty.


\lyj{{\it Surfel Management.}}  
We adopt the surfel addition and deletion \lyj{operations} proposed in \cite{yan2024gs, peng2024rtg}. During each mapping step, new surfels are added based on three criteria: (1) pixels that remain highly transparent (transmission exceeding a threshold \(\delta_T\)); (2) pixels exhibiting large depth errors (exceeding a threshold \(\delta_d\)); and (3) pixels with significant color reconstruction errors (exceeding a threshold \(\delta_c\)).  
Erroneous pixels identified by these criteria are uniformly sampled, projected \lyj{back} into 3D space, and incorporated into the surfel set.  
For surfel deletion, surfels whose accumulated average errors exceed twice the corresponding thresholds are removed in each iteration.  

\subsection{Camera Motion Tracking}
\label{tracking}

In the pose tracking \lyj{module} of \lyj{S\textsuperscript{3}LAM}, 
we utilize the analytical Jacobian of $\mathbf{SE}(3)$ derived from the backward gradient of the surfel splatting. This \lyj{algorithm} is tightly coupled with \lyj{our} model representation, \lyj{which enhances system robustness and significantly improves convergence speed.}

\begin{figure}[t]\centering
  \includegraphics[width=0.9\linewidth]{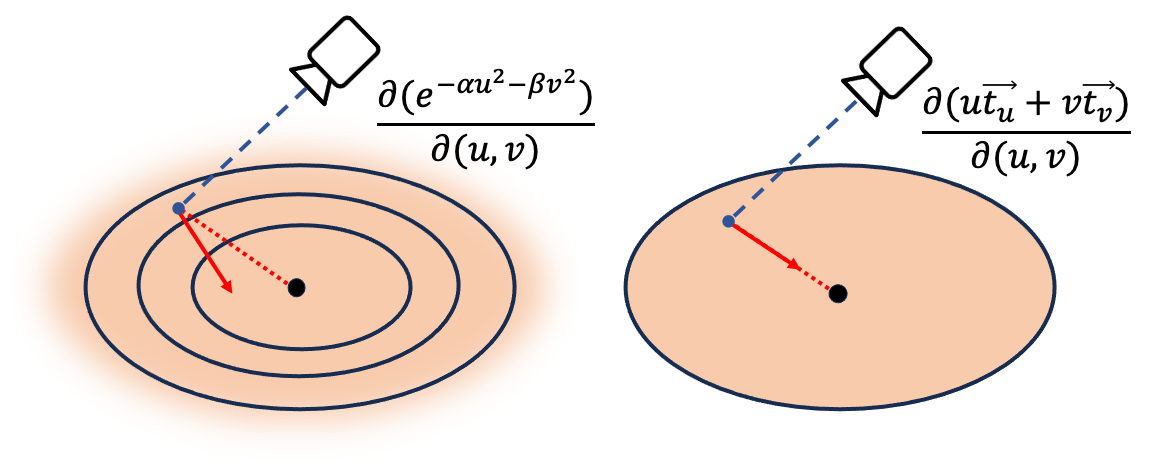}
  \caption{ \textbf{Comparison between 3D\lyj{GS}-based and 2D-\lyj{surfel}-based pose optimization.} \textbf{Left}: In 3D\lyj{GS}-based approaches, the optimization relies solely on the Jacobian \(\frac{\partial \omega}{\partial T_{CW}}\) to adjust the projected Gaussian function value, which is constrained to move perpendicular to equipotential \lyj{surfaces}. \textbf{Right}: The 2D-based Jacobian additionally introduces the radial component \(\frac{\partial \hat{\mathbf{t}}_r}{\partial T_{CW}}\) that points directly towards the surfel center.}
  \label{fig:radial}
\end{figure}

\lyj{{\it Pose Tracking with Surfel Splatting.}} 
Surfel splatting has demonstrated its capability for accurate geometric representation. Building on this, we aim to show that pose tracking can also benefit from enhanced surface accuracy. Following \cite{solà2021microlietheorystate}, we use the Lie algebra framework to derive pose Jacobians, encompassing the 3 degrees of freedom of the rotation group $\mathbf{SO}(3)$ and the 3 degrees of freedom for translation $\mathbf{R}^3$. Pose optimization is formulated as minimizing the following loss with respect to camera pose $T_{CW}$ consisting of rotation matrix $\mathbf{R}$ and translation vector $\mathbf{t}$:
\begin{small}
\begin{equation}
    L_{track} = 
    \lambda_C \left\lVert \mathbf{C}(\mathbf{S}, T_{CW}) - \bar{\mathbf{C}} \right\rVert_1 + \lambda_D\left\lVert \mathbf{D}(\mathbf{S}, T_{CW}) - \bar{\mathbf{D}} \right\rVert_1 \, .
\end{equation}
\end{small}
where $\mathbf{C}(\mathbf{S}, T_{CW})$ and $\mathbf{D}(\mathbf{S}, T_{CW})$ are the rendered image and depth map, respectively, derived from the surfel representation $\mathbf{S}$ and the estimated pose $T_{CW} \in \mathbf{SE}(3)$.

For the color loss $L_c = \left\lVert \mathbf{C}(\mathbf{S}, T_{CW}) - \bar{\mathbf{C}} \right\rVert_1$, the gradient is propagated through the surfel's central point and its two tangential vectors:


\begin{equation}
\frac{\partial L_c}{\partial T_{CW}} = \sum\limits_{\mathbf{x},\, k}\frac{\partial L_c}{\partial \mathbf{C}(\mathbf{x})}
\frac{\partial \mathbf{C}(\mathbf{x})}{\partial \omega_{k,x}}\frac{\partial \omega_{k,x}}{\partial \hat{\mathbf{q}}_k}\frac{\partial \hat{\mathbf{q}_k}}{\partial T_{CW}}
\label{equ:lc_tcw}
\end{equation}
where $\hat{\mathbf{q}}_k = (p_k, \mathbf{t}_u, \mathbf{t}_v)$ denotes the positional parameters of the k-th surfel in camera coordinates. \lyj{Most terms in our derivation align with standard 3DGS terminology, and } 
the Jacobians' \lyj{formulation} of the tangent vectors \lyj{is} derived in supplementary material. 


While the gradient of the color loss with respect to the pose is built upon the same alpha blending framework in both \lyj{3D} Gaussian spheres and \lyj{2D} surfels, their formulations for depth rendering differ significantly. In 3DGS, the depth is computed as a weighted average of the projected Gaussian centers: $\mathbf{D}(\mathbf{x}) = \sum\limits_i \omega_i\, \hat{\mathbf{p}}_{i,z}$ where \(\hat{\mathbf{p}}_{i,z}\) is the \(z\)-coordinate of the \(i\)-th Gaussian center after being transformed under the current viewpoint ~\cite{yan2024gs, matsuki2024gaussian}. \lyj{Due to the unoriented nature of Gaussian spheres}, the camera pose rotation is governed solely by the Jacobians of the alpha weights, \(\frac{\partial \omega}{\partial T_{CW}}\). These Jacobians \lyj{exhibit directional bias due to} the anisotropic \lyj{characteristics} of Gaussian \lyj{spheres}, which inherently \lyj{constrains} the camera to move perpendicular to the equipotential surfaces of the Gaussian function, as illustrated in Figure~\ref{fig:radial} left.

Unlike 3DGS, depth \lyj{in 2D surfel splatting ~\cite{huang20242d}} is computed as the blending of ray-surfel intersection points: $\mathbf{D}(\mathbf{x}) = \sum\limits_i \omega_i \hat{P}(\mathbf{u}(\mathbf{x}))_{i,\, z}$, 
\lyj{for which} the gradient is additionally influenced by the intersection-center residual:

\begin{small}
\begin{equation}
\begin{aligned}
\frac{\partial \mathbf{D}(\mathbf{x})}{\partial T_{CW}} &= \frac{\partial \mathbf{D}(\mathbf{x})}{\partial \omega_{i, x}}\frac{\partial \omega_{i, x}}{\partial T_{CW}} + \frac{\partial \mathbf{D}(\mathbf{x})}{\partial \hat{P}(u(x))}\frac{\partial \hat{P}(u(x))}{\partial T_{CW}} \\ &= \frac{\partial \mathbf{D}(\mathbf{x})}{\partial \omega_{i, x}}\frac{\partial \omega_{i, x}}{\partial T_{CW}} + \frac{\partial \mathbf{D}(\mathbf{x})}{\partial \hat{P}(u(x))} (\frac{\partial \hat{p}_i}{\partial T_{CW}} + \underbrace{\frac{\partial \hat{\mathbf{t}}_r}{\partial T_{CW}}}_{\text{radial grad}})
\end{aligned}
\end{equation}
\end{small}

\lyj{where} \(\hat{P}(u(x))\) denotes the transformed intersection point between pixel ray and surfel plane, and \(\hat{\mathbf{t}}_r = R(us_u\mathbf{t}_u + vs_v\mathbf{t}_v)\) represents the vector from surfel center to ray-surfel intersection point, satisfying the relation \(\hat{P}(u(x)) = \hat{p}_i + \hat{\mathbf{t}}_r\). Compared to the alpha-blending-style depth rendering in 3DGS, the Jacobian of surfel-based depth rendering includes an additional radial vector component, \(\hat{\mathbf{t}}_r\). This component is critical for locally optimizing the pixel ray, enabling it to move closer to or farther from the surfel center by controlling the camera's rotation, as illustrated in Figure~\ref{fig:radial} right.

From a statistical perspective, the radial gradient provided by oriented surfels acts as a guidance mechanism for aligning the reconstructed surface with the ground-truth target view. This gradient directly influences the camera's orientation, encouraging the pose to rotate either perpendicular or parallel to the underlying plane. As a result, \lyj{our} surfel-based pose optimization exhibits \lyj{remarkable tracking robustness} even \lyj{with very few} frame-to-frame overlap, demonstrating a wide convergence basin and enabling stable motion tracking under \lyj{significant} viewpoint \lyj{variations}.

\lyj{Using} the derived pose Jacobian, we iteratively optimize the camera pose as new frames \lyj{arrive, taking advantage of} the geometrically accurate representation provided by the oriented surfels. Following RTG-SLAM~\cite{peng2024rtg}, we integrate the Iterative Closest Point (ICP) method~\cite{kinectfusion} in \lyj{S\textsuperscript{3}LAM} as an optional camera tracking module for synthetic datasets, \lyj{which utilizes} high-fidelity depth \lyj{information}.


\lyj{{\it Keyframing and Optimization Strategies.}} Our keyframing strategy is based on \yh{the} camera motion\yh{~\cite{cao2018real}}.  The initial frame is added to the keyframe list, and subsequent frames are designated as keyframes, if their relative rotation exceeds \yh{a threshold} $\delta_r$ or their relative translation surpasses $\delta_t$. These newly identified keyframes are appended to the keyframe list. 

The surfel map is optimized every $M$ frames using the most recent frames. Additionally, when a new keyframe is added, the surfel map is optimized using the former keyframes. After completing the entire SLAM procedure, the surfel splatting map is further refined by optimizing with all recorded keyframes, employing ten times the number of keyframe iterations to achieve the final accuracy.

\begin{figure}[t]\centering
  \includegraphics[width=\linewidth]{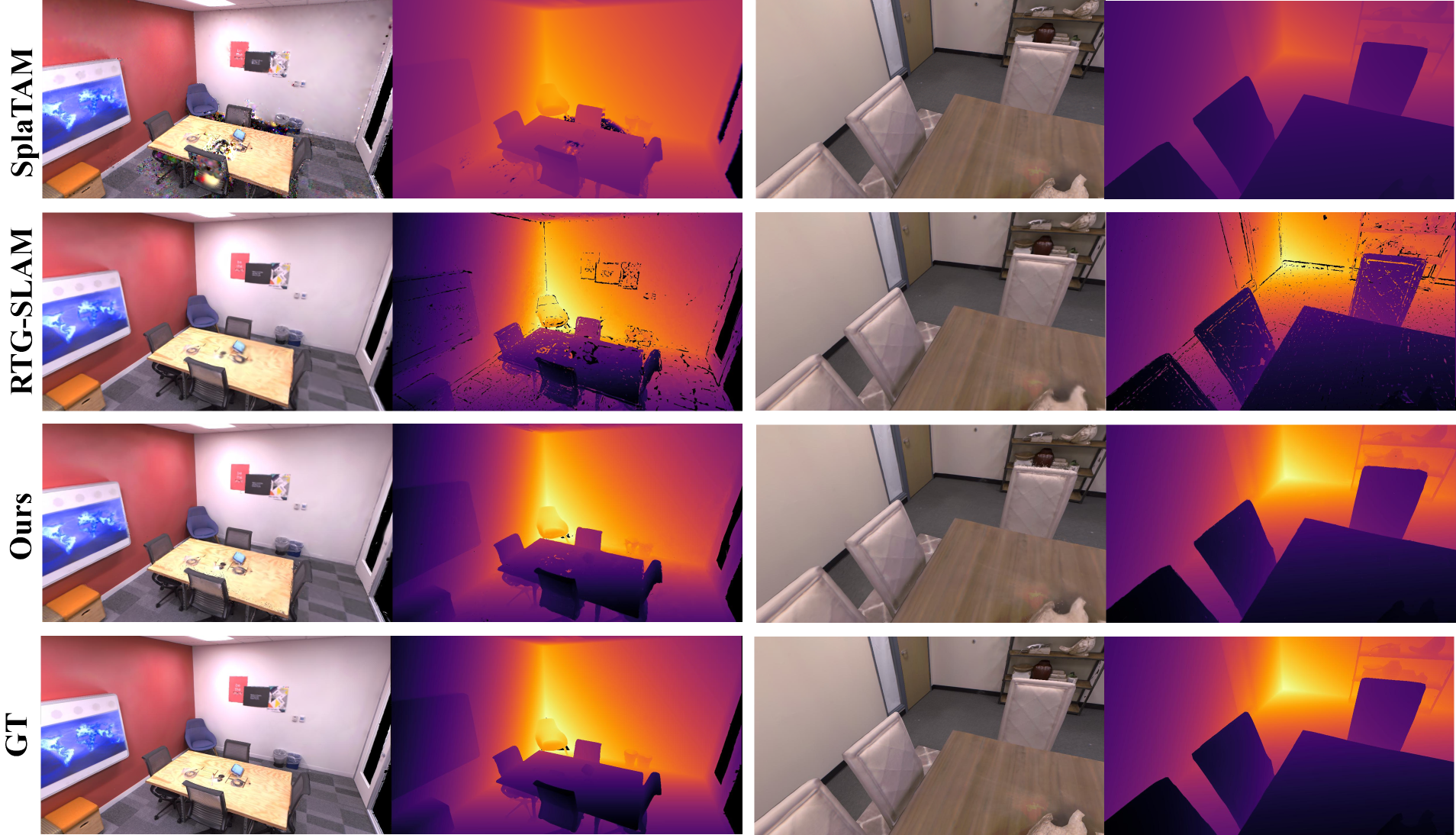}
  \caption{\textbf{Qualitative results of color and depth rendering \lyj{in} S\textsuperscript{3}LAM compared \lyj{to representative} 3DGS-based methods.} Our method exhibits fewer artifacts and clearer borders in the rendered color and depth. }
  \label{fig:qualitative}
\end{figure}

\begin{figure}[t]\centering
  \includegraphics[width=\linewidth]{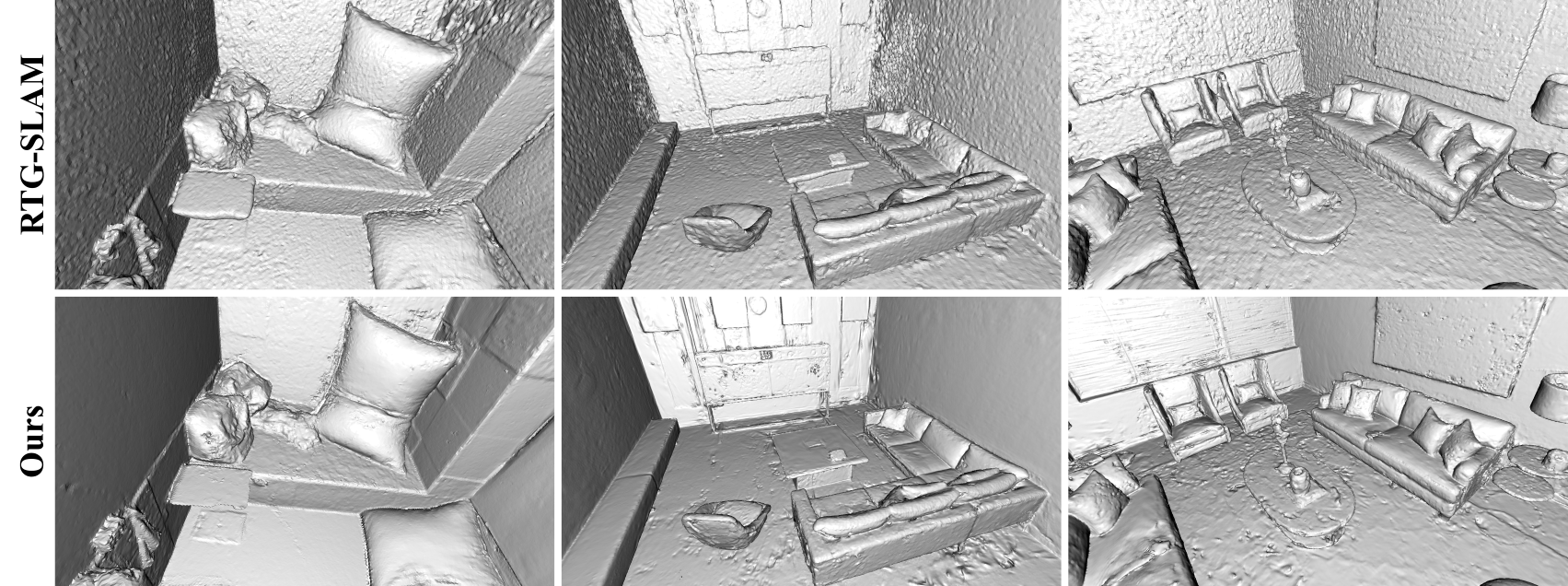}
  \caption{\textbf{Qualitative results of \lyj{3D} scene reconstruction for the comparison of S\textsuperscript{3}LAM and RTG-SLAM~\cite{peng2024rtg}.} Our method produces smoother surfaces and finer details in the reconstructed meshes.}
  \label{fig:mesh}
\end{figure}

\section{Experiments}

\subsection{Experimenal Settings}

\lyj{{\it Datasets.}}
We \lyj{conducted} experiments on three \lyj{representative} datasets: the synthetic Replica~\cite{straub2019replica}, the real-world TUM-RGBD~\cite{sturm2012benchmark} and ScanNet++~\cite{yeshwanthliu2023scannetpp}. 
\lyj{3D} reconstruction experiments are performed on the Replica dataset due to its availability of ground-truth reconstructed meshes. Pose estimation experiments are conducted on all three datasets. For ScanNet++, we selected four sequences: 8b5caf3398 (S0), b20a261fdf (S1), and f34d532901 (S2), as S0 and truncated S1 have been \lyj{previously benchmarked} in SLAM \lyj{research}~\cite{keetha2024splatam}.

\lyj{{\it Implementation Details.}} 
To implement our proposed adaptive surface mapping and pose Jacobian \lyj{algorithms}, we revised the 2D Gaussian splatting CUDA code \lyj{in} \yh{\cite{huang20242d}.} 
For the coupled \yh{pose estimation} \lyj{algorithm}, map optimization 
\yh{was} performed every 6 frames for 50 iterations, and pose optimization involved 50 iterations on Replica.
We set the mapping hyper-parameters $\gamma_D$ to 1.0, $\gamma_N$ to 0.1, $\tau$ to $5E-6$, $\delta_T$ to 0.5, $\delta_d$ and $\delta_c$ to 0.1. 
All experiments were \lyj{performed} on a Platinum 8375C CPU with \lyj{an} RTX3090 GPU. 

\lyj{{\it Baselines.}} 
We compare \lyj{the performation of S\textsuperscript{3}LAM} with (1) recent NeRF-based RGB-D SLAM methods: NICE-SLAM~\cite{zhu2022nice} and Point-SLAM~\cite{sandstrom2023point}, \lyj{and} (2) state-of-the-art 3D Gaussian splatting-based methods: Gaussian Splatting SLAM (MonoGS)~\cite{matsuki2024gaussian}, SplaTAM~\cite{keetha2024splatam}, and RTG-SLAM~\cite{peng2024rtg}.
\lyj{More} additional experimental details are \lyj{reported} in the supplementary material.

\begin{table}[tbp]
    \centering
    \caption{\textbf{Comparison of geometry accuracy on Replica.} The P and R represent precision and recall respectively. The best metric is highlighted in \textbf{bold}, and the second-best metric is \underline{underlined}.}
    \label{tab:geometry_accuracy}
    \scalebox{0.85}
    {
    \begin{tabular}{lcccccc}
        \toprule
        Method & Acc.↓ & P.↑ & Comp.↓ & R.↑ & F1↑ & L1↓\\
        \midrule
        NICE-SLAM \citeyearpar{zhu2022nice} & 2.84 & 84.4 & \underline{2.31} & 85.0 & 84.7 & 2.64 \\
        Point-SLAM \citeyearpar{sandstrom2023point} & \textbf{0.61} & \textbf{99.9} & 2.42 & \underline{86.9} & \textbf{92.7} & \textbf{0.44} \\
        \midrule
        MonoGS \citeyearpar{matsuki2024gaussian} & 2.72 & 75.6 & 3.29 & 76.6 & 75.6 & 3.56 \\
        SplaTAM \citeyearpar{keetha2024splatam}    & 2.88 & 73.9 & 3.57 & 71.7 & 72.8 & 0.70\\
        RTG-SLAM\citeyearpar{peng2024rtg}  & \underline{0.75} & \underline{98.9} & 2.81 & 82.8 & 90.0 & 1.71\\
        \textbf{Ours}              & 1.06 & 95.2 & \textbf{2.19} & \textbf{88.6} & \underline{91.9} & \underline{0.47} \\
        \bottomrule
    \end{tabular}
    }
\end{table}

\begin{table}[tbp]
    \centering
    \caption{\textbf{\lyj{Comparison on} Run-time Performance.} Metrics are obtained from run-time evaluations on the Office0 sequence of the Replica dataset.}
    \label{tab:runtime}
    \scalebox{0.8}{
    \begin{tabular}{lccc}
    \toprule
  &    FPS ↑  &  Memory ↓   &     Model Size ↓   \\
\midrule
Point-SLAM\citeyearpar{sandstrom2023point} & 0.32 &  11189 MB  &  2864 MB  \\
SplaTAM\citeyearpar{keetha2024splatam} & 0.34 & 10340 MB & 310 MB \\
MonoGS\citeyearpar{matsuki2024gaussian} & 0.87 & 25716 MB & 17.4 MB \\
RTG-SLAM\citeyearpar{peng2024rtg} & 8.95 & 3444 MB & 46.7 MB \\
\textbf{Ours} & 8.12 & 4200 MB & 43.2 MB \\
    \bottomrule
    \end{tabular}
    }
\end{table}

\subsection{Results}

\lyj{{\it 3D Scene Reconstruction.}} Following\yh{~NICE-SLAM ~\cite{zhu2022nice}}
, we evaluate \lyj{different methods} using the following metrics: accuracy (cm), precision (the proportion of reconstructed points with an accuracy below 3cm, \%), completion (cm), recall (the proportion of ground-truth points with a completion below 3cm, \%), F1-score, and L1 depth error (cm). Consistent with NeRF-based SLAM methods, we generate the scene mesh \lyj{using} the TSDF-Fusion algorithm~\cite{curless1996volumetric} after processing the RGB-D input sequence for S\textsuperscript{3}LAM and MonoGS. For RTG-SLAM and SplaTAM, we adopt their densification method to create the scene's point cloud as in \cite{peng2024rtg}.

\lyj{The} experimental results are summarized in \yh{Table~\ref{tab:geometry_accuracy}}. Our method achieves state-of-the-art performance in \lyj{3D} scene reconstruction by balancing precision and recall. 
This balance ensures that the reconstructed meshes are not only densely populated but also closely \lyj{aligned} with ground-truth vertices. 
\lyj{By combining} Gaussian surfel \lyj{representations with} our \lyj{novel} adaptive mapping strategy, \lyj{S\textsuperscript{3}LAM achieves highly accurate depth estimation from rendered maps, significantly enhancing the overall SLAM performance.}
\lyj{Quantitative evaluations in Table \ref{tab:geometry_accuracy} demonstrate that S\textsuperscript{3}LAM outperforms 3DGS-based SLAM systems while achieving reconstruction quality comparable to that of leading NeRF-based SLAMs. Notably, as shown in Table \ref{tab:runtime}, S\textsuperscript{3}LAM has substantial advantages in both computational efficiency and memory consumption compared to the NeRF-based SLAM.}

We also conduct a qualitative comparison on image rendering, depth rendering, and mesh reconstruction, as shown in Figures~\ref{fig:qualitative} and \ref{fig:mesh}. \lyj{In both rendered image and depth maps, S\textsuperscript{3}LAM generates significantly} fewer artifacts compared to \lyj{representative 3DGS-based SLAMs}. In mesh reconstruction, \lyj{the results of S\textsuperscript{3}LAM produces } more detailed object \lyj{boundaries} and smoother scene surfaces, underscoring the geometric accuracy of \lyj{S\textsuperscript{3}LAM}.

\begin{table*}[t]
\centering
\caption{\textbf{Comparison of tracking performance on Replica with ATE (unit: cm).} * denotes SLAMs utilizing ICP for tracking.}
\label{tab:replica_tracking}
\scalebox{0.9}
{
\begin{tabular}{lccccccccc}
\toprule
\textbf
{Method} & \textbf{Rm 0} & \textbf{Rm 1} & \textbf{Rm 2} & \textbf{Off 0} & \textbf{Off 1} & \textbf{Off 2} & \textbf{Off 3} & \textbf{Off 4} & \textbf{Avg.} \\ 
\midrule
NICE-SLAM\citeyearpar{zhu2022nice} & 0.97 & 1.31 & 1.07 & 0.88 & 1.00 & 1.10 & 1.10 & 1.13 & 1.06 \\
Point-SLAM\citeyearpar{sandstrom2023point} & 0.54 & 0.43 & 0.34 & 0.36 & 0.45 & 0.44 & 0.63 & 0.72 & 0.49 \\
SplaTAM\citeyearpar{keetha2024splatam}    & 0.47 & 0.42 & 0.32 & 0.46 & 0.24 & 0.28 & 0.39 & 0.56 & 0.39 \\
MonoGS\citeyearpar{matsuki2024gaussian} & 0.64 & \underline{0.34} & 0.37 & 0.47 & 0.66 & \underline{0.25} & \textbf{0.16} & 2.36 & 0.66 \\
RTG-SLAM\citeyearpar{peng2024rtg}\textsuperscript{*}    & \underline{0.20} & \textbf{0.18} & \underline{0.13} & \underline{0.22} & \textbf{0.12} & \textbf{0.22} & 0.20 & \textbf{0.19} & \textbf{0.18} \\ 
\midrule
\textbf{Ours-ICP\textsuperscript{*}}    & \textbf{{0.19}} & \textbf{{0.18}} & \textbf{0.11} & \textbf{{0.18}} & \underline{0.13} & 0.29 & \underline{0.17} & \underline{0.22} & \underline{0.21} \\ 
\textbf{Ours-Coupled}    & 0.29 & 0.68 & {0.39} & \underline{0.22} & 0.15 & {0.48} & {0.46} & {0.35} & 0.38 \\ 
\bottomrule
\end{tabular}
}
\end{table*}

\begin{table}[h]
    \centering
    \caption{\textbf{Comparison of tracking performance on TUM-RGBD with ATE (unit: cm).}}
    \label{tab:tum_rgbd}
    \scalebox{0.9}{
        \begin{tabular}{lcccc}
            \toprule
            \textbf{Methods}  & \textbf{fr1/desk} & \textbf{fr2/xyz} & \textbf{fr3/office} & \textbf{Avg.} \\
            \midrule
            NICE-SLAM \citeyearpar{zhu2022nice}    & 4.26 & 6.19 & 6.87 & 5.77 \\
            Point-SLAM \citeyearpar{sandstrom2023point}   & 4.34 & 1.31 & 3.48 & 3.04 \\
            SplaTAM  \citeyearpar{keetha2024splatam}     & 3.35 & \underline{1.24} & 5.16 & 3.25 \\
            MonoGS  \citeyearpar{matsuki2024gaussian}      & \textbf{1.52} & 1.58 & \textbf{1.65} & \textbf{1.58} \\
            \textbf{Ours}  & \underline{2.38} & \textbf{1.16} & \underline{2.26} & \underline{1.93} \\
            \bottomrule
        \end{tabular}
    }
\end{table}

\begin{table}[h]
    \centering
    \caption{\textbf{Comparison of tracking performance on ScanNet++ with ATE (unit: cm).} The S1* denotes the truncated sequence for S1.} 
    \label{tab:scannetpp}
    \scalebox{0.9}{
        \begin{tabular}{lcccc}
            \toprule
            \textbf{Methods} & \textbf{S0} & \textbf{S1*} & \textbf{S1} & \textbf{S2} \\
            \midrule
            Point-SLAM\citeyearpar{sandstrom2023point} & 411.9  & 1000.8 & 829.4 & 1888.4 \\
            MonoGS\citeyearpar{matsuki2024gaussian}     & 141.5 & 106.9 & 122.7 & 147.7 \\
            SplaTAM\citeyearpar{keetha2024splatam}     & \underline{0.63} & \underline{1.90} & \underline{70.93} & \underline{2.05} \\
            \textbf{Ours} & \textbf{0.35} & \textbf{0.42} & \textbf{0.51} & \textbf{1.11} \\
            \bottomrule
        \end{tabular}
    }
\end{table}

\lyj{{\it Pose Tracking.}}
We evaluate \lyj{baseline methods} and our \lyj{two} pose tracking strategies on the Replica dataset. The experimental results are \lyj{summarized} in Table \ref{tab:replica_tracking}. \lyj{In S\textsuperscript{3}LAM,} we refer to the method {\it pose tracking with surfel splatting} and the ICP-based tracking as \textbf{Ours-Coupled} and \textbf{Ours-ICP}, respectively. 
On the Replica dataset, due to high-quality depth data and \lyj{small} pose variations, the ICP method achieves \lyj{very good} performance. Compared to other coupled methods using Gaussian gradients, our surfel-based tracking method demonstrates superior performance.

For the TUM-RGBD and ScanNet++ datasets, where depth images are noisy and inter-frame motions can be large, we use only the coupled surfel-based tracking without ICP, as it offers better convergence and smaller drift.
The results on the TUM\yh{-RGBD} dataset are shown in Table \ref{tab:tum_rgbd}. It is worth noting that the significant noise and incompleteness of depth data in TUM-RGBD have \lyj{small} impact on MonoGS~\citeyearpar{matsuki2024gaussian}, which \lyj{primarily} relies on color information. \lyj{On the other hand}, our method outperforms other NeRF-based and 3DGS-based RGB-D SLAM methods, and achieves \lyj{competitive} performance with MonoGS~\citeyearpar{matsuki2024gaussian}.


The tracking results \lyj{on} the ScanNet++ dataset are shown in Table \ref{tab:scannetpp}. ScanNet++ features abrupt teleportations and rotations in its trajectories, causing many methods to either fail to track camera motion or track only partial \lyj{camera} motion.
SplaTAM~\cite{keetha2024splatam} was originally tested on S0 and truncated S1 from 
the 0-th frame to the 360-th frame, due to intense relative movement in the subsequent frames. 

In contrast, our method demonstrates robust tracking convergence even when adjacent frames exhibit \lyj{very small} overlap, achieving state-of-the-art performance. For example, in S1 between frames 363 and 364, where a significant rotation causes other methods to fail, our method successfully tracks the large viewpoint change. 
Thanks to our proposed surfel-based pose optimization, \lyj{S\textsuperscript{3}LAM} successfully converges to the correct pose, 
which is achieved through the Jacobian formulation of the oriented surfels, as derived in our pose tracking method. \lyj{In the next section}, a \lyj{detailed} analysis of the tracking convergence is provided.

\begin{table}[h]
    \centering
    \caption{\textbf{Ablation study for adaptive mapping.}} 
    
    \label{tab:ablation}
    \scalebox{0.85}{
        \begin{tabular}{lcccccc}
            \toprule
             & Acc.↓ & P.↑ & Comp.↓ & R.↑ & F1↑ & L1↓\\
            \midrule
            3DGS & 2.80 & 3.71 & 74.8 & 70.1 & 72.3 & 2.21 \\
            Mean   & 1.41 & 91.5 & \textbf{2.09} & \textbf{88.9} & 90.1 & \underline{0.53} \\
            Median   & \underline{1.10} & \underline{94.7} & 2.35 & 87.1 & \underline{90.6} & 0.71 \\
            \textbf{Adaptive}    & \textbf{1.06} & \textbf{95.2} & \underline{2.19} & \underline{88.6} & \textbf{91.9} & \textbf{0.47} \\
            \bottomrule
        \end{tabular}
    }
\end{table}

\begin{table}[h]
    \centering
    \caption{\textbf{Ablation study comparing tracking accuracy with ATE (unit: cm) on ScanNet++.}} 
    \label{tab:ablation_track}
    \scalebox{0.9}{
        \begin{tabular}{lcccccc}
        \toprule
         &	\textbf{S0} & 	\textbf{S1*} & 	\textbf{S1} & 	\textbf{S2} \\
         \midrule
        w/o depth loss & 	162.5 &	138.5 &	168.4 &	150.1 \\
        w/o radial &	224.0 &	331.4 &	567.2 &	200.8 \\
        \textbf{Ours} &	\textbf{0.68} &	\textbf{1.21} &	\textbf{1.43} &	\textbf{1.69} \\
        \bottomrule
        \end{tabular}
    }
\end{table}

\begin{figure}[t]\centering
  \includegraphics[width=\linewidth]{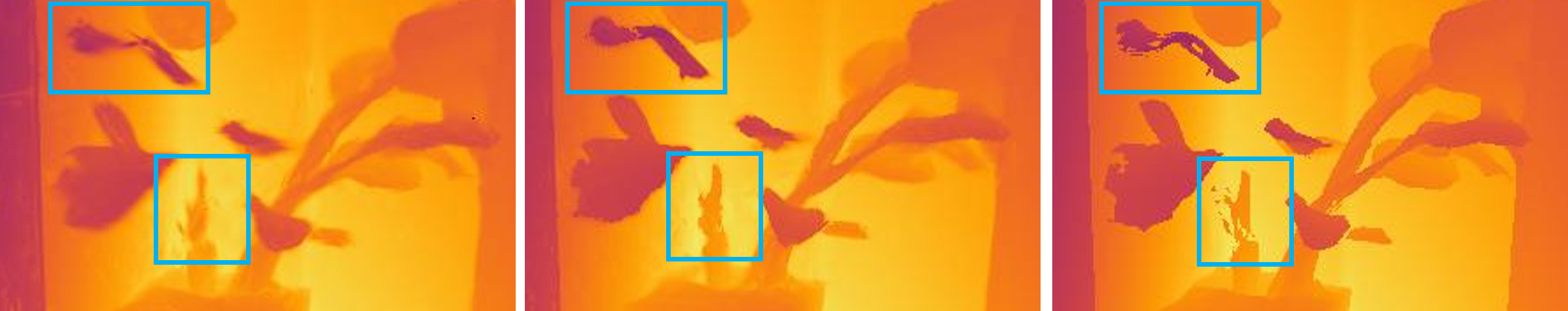}
  \caption{\textbf{Qualitative ablation results:} Depth reconstruction without the adaptive strategy using mean depth  (\textbf{Left}), with our proposed distortion-based adaptive strategy (\textbf{middle}), and the ground-truth depth (\textbf{right}).}
  \label{fig:ablation}
\end{figure}

\subsection{Convergence Basin Analysis}

\begin{figure}[t]
\centering
\begin{minipage}{0.48\linewidth}
\centering
  \includegraphics[width=\linewidth]{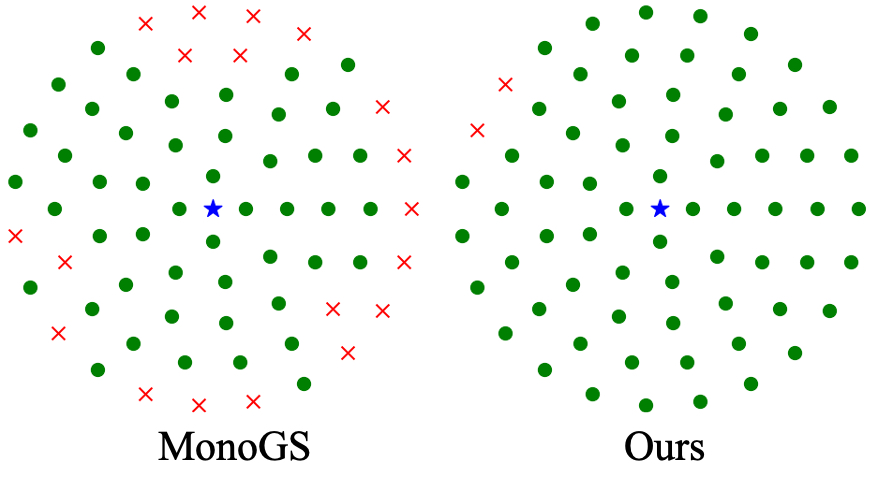}
\end{minipage}
\hfill
\begin{minipage}{0.48\linewidth}
    \centering
    \label{tab:pose}
    \scalebox{0.75}{
    \begin{tabular}{lccc}
        \toprule
         & Seq1 & Seq2 & Seq3 \\
        \midrule
        MonoGS  & 0.72 & 0.46 & 0.55\\
        \textbf{Ours}   & \textbf{0.97} & \textbf{0.77} & \textbf{1.00}\\
        \bottomrule
    \end{tabular}
    }
\end{minipage}
\caption{\textbf{\lyj{Comparison on} convergence basin (left) and tracking converging rate (right).}}
  \label{fig:cba}
\end{figure}

    



We follow \cite{mitra2004registration, matsuki2024gaussian} to evaluate pose convergence.
To simulate SLAM’s limited perception, we train the scene using only a restricted 3×3 grid of views with 0.1m spacing. We sample initial camera positions from 0.2m to 1.6m from the target view (center view of training data) and run 1,000 pose optimization steps using the RGB and depth of the target view. We count the run as successful if the final pose is within 1cm of the target.

We compare our method with a \lyj{representative 3DGS}-based pose optimization~\cite{matsuki2024gaussian} on three Replica sequences. 
Figure~\ref{fig:cba} shows quantitative success rates\footnote{Since MonoGS~\cite{matsuki2024gaussian} does not specify exact sequences, our selected sequences may yield different results.} and the visualized convergence basins. 
Our \lyj{method} achieves better convergence even when initial poses are far from the target, with \lyj{very small} view overlap. This improved behavior explains the strong tracking performance on ScanNet++, indicating that pose tracking with oriented surfel primitives significantly expands the convergence basin under large viewpoint changes.

\subsection{Ablation Study}
We conducted an ablation study to evaluate the impact of distortion-adaptive rendering. The quantitative results are presented in Table~\ref{tab:ablation}, comparing our method with 3D Gaussian splatting, 2D Gaussian splatting reconstructions based on mean depth and median depth strategies ~\cite{huang20242d}. 
Mapping with 3D Gaussian primitives yields suboptimal depth estimation and reconstruction performance. Similarly, without using depth distortion as a measure of mapping uncertainty, surfel splatting mapping also results in suboptimal reconstruction.
Interestingly, reconstruction based on mean depth achieves higher recall, while median depth provides better precision. Our method, which combines average depth blending with dominant surfel-determined depth rendering, successfully captures the benefits of both strategies, achieving comparable precision and superior recall. 
Qualitative results shown in Figure~\ref{fig:ablation} further highlight the degradation in reconstruction quality when distortion awareness is omitted, underscoring the importance of incorporating depth distortion into the mapping process.

For the ablation of pose optimization, we evaluate performance without the depth tracking loss and with the radial Jacobian explicitly removed from the pose Jacobian on ScanNet++. The results in Table~\ref{tab:ablation_track} validate that the \lyj{whole} pose tracking design leads to better convergence and accuracy.

\section{Conclusion}

In this paper, we \lyj{propose} S\textsuperscript{3}LAM, a novel SLAM system that leverages \lyj{2D} surfel splatting as its core representation for mapping. By utilizing oriented 2D Gaussian surfels, our method achieves superior surface reconstruction accuracy and robust pose tracking capability, even in challenging scenarios. A current limitation is \lyj{its degraded} performance with low-quality depth sensor inputs. In future work, we plan to enhance our surfel-based mapping and tracking under noisy sensor conditions.

\bibliography{aaai2026}

\begin{thebibliography}{32}
\providecommand{\natexlab}[1]{#1}

\bibitem[{Barron et~al.(2022)Barron, Mildenhall, Verbin, Srinivasan, and Hedman}]{Barron_Mip_NeRF}
Barron, J.~T.; Mildenhall, B.; Verbin, D.; Srinivasan, P.~P.; and Hedman, P. 2022.
\newblock Mip-NeRF 360: Unbounded Anti-Aliased Neural Radiance Fields.
\newblock In \emph{Proceedings of the IEEE/CVF Conference on Computer Vision and Pattern Recognition (CVPR)}, 5470--5479. IEEE.

\bibitem[{Cao, Kobbelt, and Hu(2018)}]{cao2018real}
Cao, Y.-P.; Kobbelt, L.; and Hu, S.-M. 2018.
\newblock Real-time high-accuracy three-dimensional reconstruction with consumer RGB-D cameras.
\newblock \emph{ACM Transactions on Graphics (TOG)}, 37(5): 1--16.

\bibitem[{Curless and Levoy(1996)}]{curless1996volumetric}
Curless, B.; and Levoy, M. 1996.
\newblock A volumetric method for building complex models from range images.
\newblock In \emph{Proceedings of the 23rd annual conference on Computer graphics and interactive techniques}, 303--312.

\bibitem[{Dai et~al.(2017)Dai, Nie{\ss}ner, Zoll{\"o}fer, Izadi, and Theobalt}]{dai2017bundlefusion}
Dai, A.; Nie{\ss}ner, M.; Zoll{\"o}fer, M.; Izadi, S.; and Theobalt, C. 2017.
\newblock BundleFusion: Real-time Globally Consistent 3D Reconstruction using On-the-fly Surface Re-integration.
\newblock \emph{ACM Transactions on Graphics 2017 (TOG)}.

\bibitem[{Dai et~al.(2024)Dai, Xu, Xie, Liu, Wang, and Xu}]{dai2024high}
Dai, P.; Xu, J.; Xie, W.; Liu, X.; Wang, H.; and Xu, W. 2024.
\newblock High-quality surface reconstruction using gaussian surfels.
\newblock In \emph{ACM SIGGRAPH 2024 Conference Papers}, 1--11. ACM.

\bibitem[{Davison et~al.(2007)Davison, Reid, Molton, and Stasse}]{monoslam}
Davison, A.~J.; Reid, I.~D.; Molton, N.~D.; and Stasse, O. 2007.
\newblock MonoSLAM: Real-Time Single Camera SLAM.
\newblock \emph{IEEE Transactions on Pattern Analysis and Machine Intelligence}, 29(6): 1052--1067.

\bibitem[{Hu et~al.(2025)Hu, Chen, Feng, Li, Yang, Bao, Zhang, and Cui}]{hu2025cg}
Hu, J.; Chen, X.; Feng, B.; Li, G.; Yang, L.; Bao, H.; Zhang, G.; and Cui, Z. 2025.
\newblock Cg-slam: Efficient dense rgb-d slam in a consistent uncertainty-aware 3d gaussian field.
\newblock In \emph{European Conference on Computer Vision}, 93--112. Springer, Springer.

\bibitem[{Huang et~al.(2024)Huang, Yu, Chen, Geiger, and Gao}]{huang20242d}
Huang, B.; Yu, Z.; Chen, A.; Geiger, A.; and Gao, S. 2024.
\newblock 2d gaussian splatting for geometrically accurate radiance fields.
\newblock In \emph{ACM SIGGRAPH 2024 conference papers}, 1--11. ACM.

\bibitem[{Huang et~al.(2021)Huang, Huang, Song, and Hu}]{huang2021di}
Huang, J.; Huang, S.-S.; Song, H.; and Hu, S.-M. 2021.
\newblock Di-fusion: Online implicit 3d reconstruction with deep priors.
\newblock In \emph{Proceedings of the IEEE/CVF Conference on Computer Vision and Pattern Recognition}, 8932--8941. Computer Vision Foundation / IEEE.

\bibitem[{Keetha et~al.(2024)Keetha, Karhade, Jatavallabhula, Yang, Scherer, Ramanan, and Luiten}]{keetha2024splatam}
Keetha, N.; Karhade, J.; Jatavallabhula, K.~M.; Yang, G.; Scherer, S.; Ramanan, D.; and Luiten, J. 2024.
\newblock SplaTAM: Splat Track \& Map 3D Gaussians for Dense RGB-D SLAM.
\newblock In \emph{Proceedings of the IEEE/CVF Conference on Computer Vision and Pattern Recognition}, 21357--21366. IEEE.

\bibitem[{Kerbl et~al.(2023)Kerbl, Kopanas, Leimk{\"u}hler, and Drettakis}]{kerbl3Dgaussians}
Kerbl, B.; Kopanas, G.; Leimk{\"u}hler, T.; and Drettakis, G. 2023.
\newblock 3D Gaussian Splatting for Real-Time Radiance Field Rendering.
\newblock \emph{ACM Transactions on Graphics}, 42(4): 139--1.

\bibitem[{Matsuki et~al.(2024)Matsuki, Murai, Kelly, and Davison}]{matsuki2024gaussian}
Matsuki, H.; Murai, R.; Kelly, P.~H.; and Davison, A.~J. 2024.
\newblock Gaussian splatting slam.
\newblock In \emph{Proceedings of the IEEE/CVF Conference on Computer Vision and Pattern Recognition}, 18039--18048.

\bibitem[{Mildenhall et~al.(2021)Mildenhall, Srinivasan, Tancik, Barron, Ramamoorthi, and Ng}]{mildenhall2021nerf}
Mildenhall, B.; Srinivasan, P.~P.; Tancik, M.; Barron, J.~T.; Ramamoorthi, R.; and Ng, R. 2021.
\newblock Nerf: Representing scenes as neural radiance fields for view synthesis.
\newblock \emph{Communications of the ACM}, 65(1): 99--106.

\bibitem[{Mitra et~al.(2004)Mitra, Gelfand, Pottmann, and Guibas}]{mitra2004registration}
Mitra, N.~J.; Gelfand, N.; Pottmann, H.; and Guibas, L. 2004.
\newblock Registration of point cloud data from a geometric optimization perspective.
\newblock In \emph{Proceedings of the 2004 Eurographics/ACM SIGGRAPH symposium on Geometry processing}, 22--31.

\bibitem[{Newcombe et~al.(2011)Newcombe, Izadi, Hilliges, Molyneaux, Kim, Davison, Kohi, Shotton, Hodges, and Fitzgibbon}]{kinectfusion}
Newcombe, R.~A.; Izadi, S.; Hilliges, O.; Molyneaux, D.; Kim, D.; Davison, A.~J.; Kohi, P.; Shotton, J.; Hodges, S.; and Fitzgibbon, A. 2011.
\newblock KinectFusion: Real-time dense surface mapping and tracking.
\newblock In \emph{2011 10th IEEE International Symposium on Mixed and Augmented Reality}, 127--136.

\bibitem[{Newcombe, Lovegrove, and Davison(2011)}]{newcombe2011dtam}
Newcombe, R.~A.; Lovegrove, S.~J.; and Davison, A.~J. 2011.
\newblock DTAM: Dense tracking and mapping in real-time.
\newblock In \emph{2011 international conference on computer vision}, 2320--2327. IEEE, IEEE.

\bibitem[{Peng et~al.(2024)Peng, Shao, Liu, Zhou, Yang, Wang, and Zhou}]{peng2024rtg}
Peng, Z.; Shao, T.; Liu, Y.; Zhou, J.; Yang, Y.; Wang, J.; and Zhou, K. 2024.
\newblock Rtg-slam: Real-time 3d reconstruction at scale using gaussian splatting.
\newblock In \emph{ACM SIGGRAPH 2024 Conference Papers}, 1--11.

\bibitem[{Ruetz et~al.(2019)Ruetz, Hernández, Pfeiffer, Oleynikova, Cox, Lowe, and Borges}]{ovpcmesh}
Ruetz, F.; Hernández, E.; Pfeiffer, M.; Oleynikova, H.; Cox, M.; Lowe, T.; and Borges, P. 2019.
\newblock OVPC Mesh: 3D Free-space Representation for Local Ground Vehicle Navigation.
\newblock In \emph{2019 International Conference on Robotics and Automation (ICRA)}, 8648--8654. IEEE.

\bibitem[{Sandstr{\"o}m et~al.(2023)Sandstr{\"o}m, Li, Van~Gool, and Oswald}]{sandstrom2023point}
Sandstr{\"o}m, E.; Li, Y.; Van~Gool, L.; and Oswald, M.~R. 2023.
\newblock Point-slam: Dense neural point cloud-based slam.
\newblock In \emph{Proceedings of the IEEE/CVF International Conference on Computer Vision}, 18433--18444. IEEE.

\bibitem[{Solà, Deray, and Atchuthan(2021)}]{solà2021microlietheorystate}
Solà, J.; Deray, J.; and Atchuthan, D. 2021.
\newblock A micro Lie theory for state estimation in robotics.
\newblock arXiv:1812.01537.

\bibitem[{Straub et~al.(2019)Straub, Whelan, Ma, Chen, Wijmans, Green, Engel, Mur-Artal, Ren, Verma et~al.}]{straub2019replica}
Straub, J.; Whelan, T.; Ma, L.; Chen, Y.; Wijmans, E.; Green, S.; Engel, J.~J.; Mur-Artal, R.; Ren, C.; Verma, S.; et~al. 2019.
\newblock The Replica dataset: A digital replica of indoor spaces.
\newblock \emph{arXiv preprint arXiv:1906.05797}.

\bibitem[{Sturm et~al.(2012)Sturm, Engelhard, Endres, Burgard, and Cremers}]{sturm2012benchmark}
Sturm, J.; Engelhard, N.; Endres, F.; Burgard, W.; and Cremers, D. 2012.
\newblock A benchmark for the evaluation of RGB-D SLAM systems.
\newblock In \emph{2012 IEEE/RSJ international conference on intelligent robots and systems}, 573--580. IEEE, IEEE.

\bibitem[{Su et~al.(2025)Su, Chen, Zhang, Zhao, Hou, and Yu}]{su2025gausslamdensergbdslam}
Su, Y.; Chen, L.; Zhang, K.; Zhao, Z.; Hou, C.; and Yu, Z. 2025.
\newblock GauS-SLAM: Dense RGB-D SLAM with Gaussian Surfels.
\newblock arXiv:2505.01934.

\bibitem[{Sucar et~al.(2021)Sucar, Liu, Ortiz, and Davison}]{sucar2021imap}
Sucar, E.; Liu, S.; Ortiz, J.; and Davison, A.~J. 2021.
\newblock imap: Implicit mapping and positioning in real-time.
\newblock In \emph{Proceedings of the IEEE/CVF international conference on computer vision}, 6229--6238. IEEE.

\bibitem[{Sucar, Wada, and Davison(2020)}]{sucar2020nodeslam}
Sucar, E.; Wada, K.; and Davison, A. 2020.
\newblock NodeSLAM: Neural object descriptors for multi-view shape reconstruction.
\newblock In \emph{2020 International Conference on 3D Vision (3DV)}, 949--958. IEEE.

\bibitem[{Teed and Deng(2021)}]{teed2021droid}
Teed, Z.; and Deng, J. 2021.
\newblock Droid-slam: Deep visual slam for monocular, stereo, and rgb-d cameras.
\newblock \emph{Advances in neural information processing systems}, 34: 16558--16569.

\bibitem[{Whelan et~al.(2016)Whelan, Salas-Moreno, Glocker, Davison, and Leutenegger}]{whelan2016elasticfusion}
Whelan, T.; Salas-Moreno, R.~F.; Glocker, B.; Davison, A.~J.; and Leutenegger, S. 2016.
\newblock ElasticFusion: Real-time dense SLAM and light source estimation.
\newblock \emph{The International Journal of Robotics Research}, 35(14): 1697--1716.

\bibitem[{Yan et~al.(2024)Yan, Qu, Xu, Zhao, Wang, Wang, and Li}]{yan2024gs}
Yan, C.; Qu, D.; Xu, D.; Zhao, B.; Wang, Z.; Wang, D.; and Li, X. 2024.
\newblock Gs-slam: Dense visual slam with 3d gaussian splatting.
\newblock In \emph{Proceedings of the IEEE/CVF Conference on Computer Vision and Pattern Recognition}, 19595--19604. IEEE.

\bibitem[{Yang et~al.(2022)Yang, Li, Zhai, Ming, Liu, and Zhang}]{yang2022vox}
Yang, X.; Li, H.; Zhai, H.; Ming, Y.; Liu, Y.; and Zhang, G. 2022.
\newblock Vox-Fusion: Dense Tracking and Mapping with Voxel-based Neural Implicit Representation.
\newblock In \emph{2022 IEEE International Symposium on Mixed and Augmented Reality (ISMAR)}, 499--507. IEEE.

\bibitem[{Yeshwanth et~al.(2023)Yeshwanth, Liu, Nie{\ss}ner, and Dai}]{yeshwanthliu2023scannetpp}
Yeshwanth, C.; Liu, Y.-C.; Nie{\ss}ner, M.; and Dai, A. 2023.
\newblock ScanNet++: A High-Fidelity Dataset of 3D Indoor Scenes.
\newblock In \emph{Proceedings of the International Conference on Computer Vision ({ICCV})}, 12--22. IEEE.

\bibitem[{Zhi et~al.(2019)Zhi, Bloesch, Leutenegger, and Davison}]{zhi2019scenecode}
Zhi, S.; Bloesch, M.; Leutenegger, S.; and Davison, A.~J. 2019.
\newblock Scenecode: Monocular dense semantic reconstruction using learned encoded scene representations.
\newblock In \emph{Proceedings of the IEEE/CVF Conference on Computer Vision and Pattern Recognition}, 11776--11785. IEEE.

\bibitem[{Zhu et~al.(2022)Zhu, Peng, Larsson, Xu, Bao, Cui, Oswald, and Pollefeys}]{zhu2022nice}
Zhu, Z.; Peng, S.; Larsson, V.; Xu, W.; Bao, H.; Cui, Z.; Oswald, M.~R.; and Pollefeys, M. 2022.
\newblock Nice-slam: Neural implicit scalable encoding for slam.
\newblock In \emph{Proceedings of the IEEE/CVF conference on computer vision and pattern recognition}, 12786--12796. IEEE.

\end{thebibliography}

\end{document}